%% file: main.tex
\documentclass[10pt,twocolumn,letterpaper]{article}

\usepackage{cvpr}
\usepackage{times}
\usepackage{epsfig}
\usepackage{graphicx}
\usepackage{amsmath}
\usepackage{amssymb}
\usepackage{mathtools}
\usepackage{verbatim}
\usepackage{caption}
\usepackage{subcaption}
\usepackage{tabularx}
\usepackage{colortbl}
\usepackage{etoolbox}
\usepackage{booktabs}
\usepackage{arydshln}
\usepackage{makecell}
\usepackage{colortbl}
\usepackage{tabu}
\usepackage{multirow}
\usepackage{enumitem}
\usepackage{setspace}
\usepackage[dvipsnames]{xcolor}
\usepackage[skip=1ex,font=small,labelsep=period]{caption}
\usepackage[pagebackref=true,breaklinks=true,letterpaper=true,colorlinks=true,bookmarks=false,allcolors=black]{hyperref}

\definecolor{bblue}{rgb}{0.0,0.25,0.65}
\definecolor{ccol}{rgb}{0.2,0.2,0.2}

\newcolumntype{Y}{>{\centering\arraybackslash}X}
\newcolumntype{R}{>{\raggedleft\arraybackslash}X}
\newcolumntype{L}{>{\raggedright\arraybackslash}X}

\newcommand\customparagraph[1]{\vspace{0.7em}\noindent\textbf{#1}}

\newcommand{\win}[1]{\textcolor{NavyBlue}{\textbf{#1}}}
\newcommand{\wincat}[1]{\textcolor{black}{\textbf{#1}}}
\newcommand\myeq{\mkern1.5mu{=}\mkern1.5mu}

\definecolor{lightgray}{rgb}{0.935, 0.935, 0.935}

\makeatletter
\def\adl@drawiv#1#2#3{%
        \hskip.5\tabcolsep
        \xleaders#3{#2.5\@tempdimb #1{1}#2.5\@tempdimb}%
                #2\z@ plus1fil minus1fil\relax
        \hskip.5\tabcolsep}
\newcommand{\cdashlinelr}[1]{%
  \noalign{\vskip\aboverulesep
           \global\let\@dashdrawstore\adl@draw
           \global\let\adl@draw\adl@drawiv}
  \cdashline{#1}
  \noalign{\global\let\adl@draw\@dashdrawstore
           \vskip\belowrulesep}}
\makeatother

\makeatletter
\newlength{\qrr@dimen@}
\expandafter\pretocmd\csname tabular*\endcsname{\setlength{\qrr@dimen@}{#1}}{}{}
\newcommand*{\Rowcolor}[2][\tabcolsep]{%
    \ifx\relax#1\relax\else
        \kern-\the\dimexpr#1\relax
    \fi
    \makebox[0pt][l]{%
        \fboxsep=0pt
        \colorbox{#2}{%
            \strut\kern\qrr@dimen@
        }%
    }%
    \ifx\relax#1\relax\else
        \kern\the\dimexpr#1\relax
    \fi
    \ignorespaces
}
\makeatother

\cvprfinalcopy % *** Uncomment this line for the final submission

\def\cvprPaperID{5280} % *** Enter the CVPR Paper ID here

\begin{document}

\title{EPOS: Estimating 6D Pose of Objects with Symmetries}

\newcommand{\namesep}{\hspace{1.0em}}
\author{
 Tom{\'a}{\v{s}}~Hoda{\v{n}}$^{1}$\namesep
 D{\'a}niel~Bar{\'a}th$^{1,2}$\namesep
 Ji{\v{r}}{\'i}~Matas$^{1}$ \vspace{0.7em} \\
 {$^{1}$Visual Recognition Group, Czech Technical University in Prague} \\
 {$^{2}$Machine Perception Research Laboratory, MTA SZTAKI, Budapest} \\
}

\maketitle
%\ifcvprfinal\thispagestyle{empty}\fi

\begin{abstract}
We present a new method for estimating the 6D pose of rigid objects with available 3D models from a single RGB input image.
The method is applicable to a broad range of objects, including challenging ones with global or partial symmetries.
An object is represented by compact	surface fragments which allow handling symmetries in a systematic manner.
Correspondences between densely sampled pixels and the fragments are predicted using an encoder-decoder network.
At each pixel, the network predicts: (i)~the probability of each object's presence, (ii)~the probability of the fragments given the object's presence, and (iii)~the precise 3D location on each fragment.
A data-dependent number of corresponding 3D locations is selected per pixel, and poses of possibly multiple object instances are estimated using a robust and efficient variant of the PnP-RANSAC algorithm.
In the BOP Challenge 2019, the method outperforms all RGB and most RGB-D and D methods on the T-LESS and LM-O datasets.
On the YCB-V dataset, it is superior to all competitors, with a large margin over the second-best RGB method.
Source code is at: \texttt{\href{http://cmp.felk.cvut.cz/epos/}{cmp.felk.cvut.cz/epos}}.
\end{abstract}

%%%%%%%%%%%%%%%%%%%%%%%%%%%%%%%%%%%%%%%%%%%%%%%%%%%%%%%%%%%%%%%%%%%%%%%%%%%%%%%%
\vspace{-1ex}
\section{Introduction} \label{sec:intro}

Model-based estimation of 6D pose, \ie the 3D translation and 3D rotation, of rigid objects from a single image is a classical computer vision problem, with the first methods dating back to the work of Roberts from 1963~\cite{roberts1963machine}.
A~common approach to the problem is to establish a set of 2D-3D correspondences between the input image and the object model and robustly estimate the pose by the P\emph{n}P-RANSAC algorithm~\cite{fischler1981random,lepetit2009epnp}.
Traditional methods~\cite{collet2011moped} establish the correspondences using local image features, such as SIFT~\cite{lowe1999object}, and have demonstrated robustness against occlusion and clutter in the case of objects with distinct and non-repeatable shape or texture.
Recent methods, which are mostly based on convolutional neural networks, produce dense correspondences~\cite{brachmann2014learning,park2019pix2pose,zakharov2019dpod} or predict 2D image locations of pre-selected 3D keypoints~\cite{rad2017bb8,tekin2018real,peng2019pvnet}.

\begin{figure}[t!]
	\begin{center}
	    \begingroup
        \setlength{\tabcolsep}{0.0pt}
        \renewcommand{\arraystretch}{0}
		\begin{tabular}{ c c }
			\includegraphics[width=0.492\linewidth]{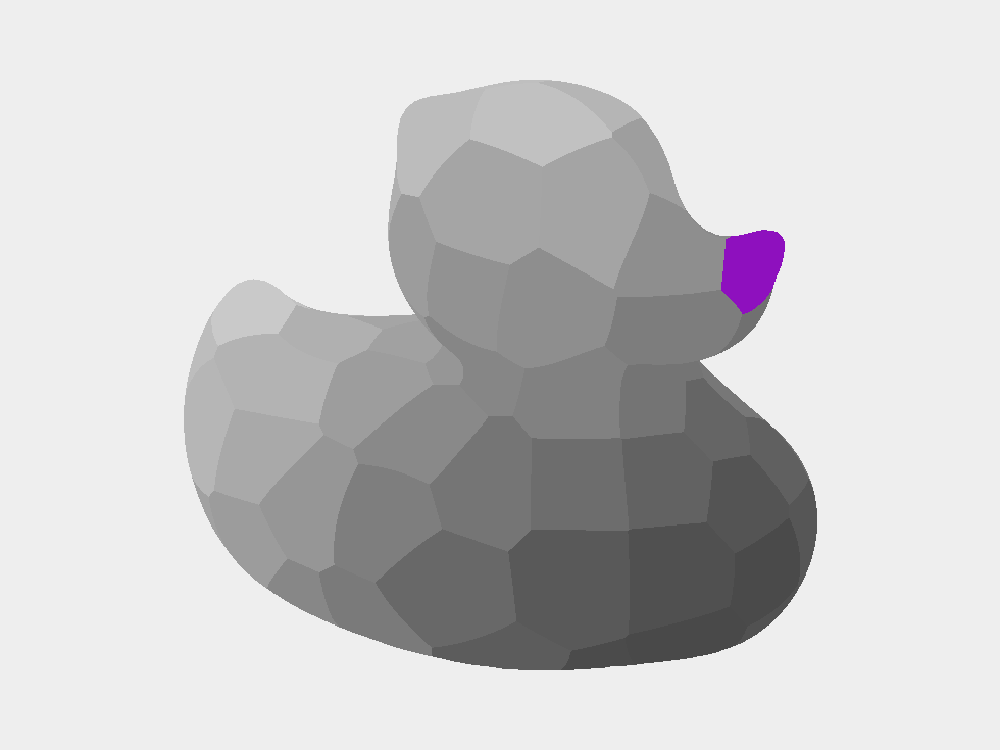} \hspace{0.25ex} &
			\includegraphics[width=0.492\linewidth]{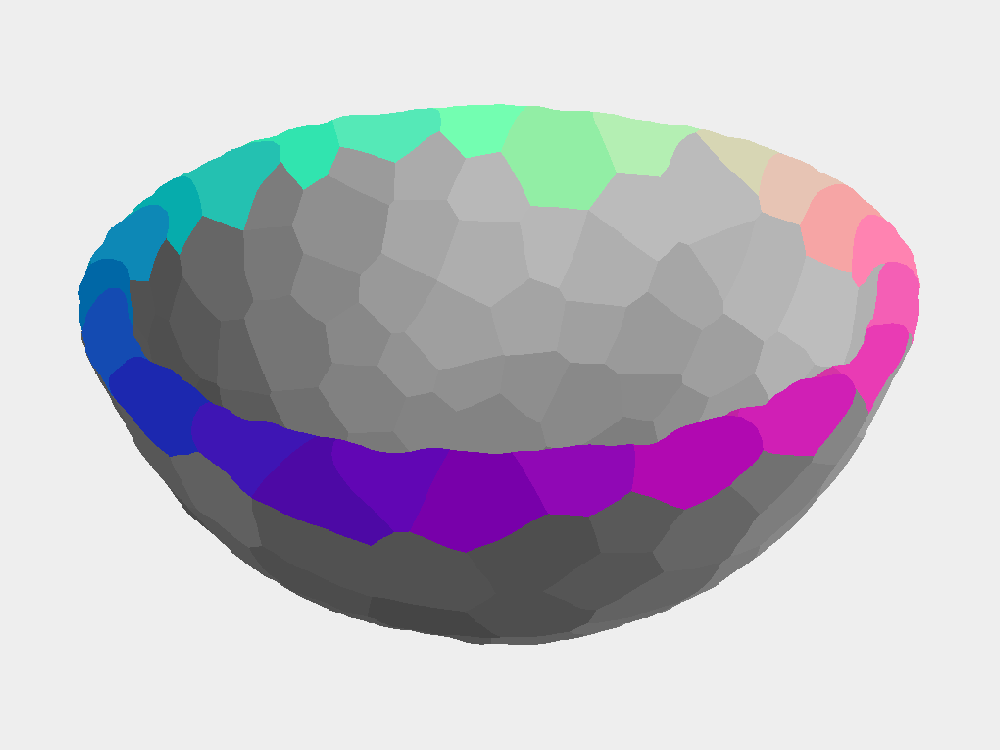} \\
			\includegraphics[width=0.492\linewidth]{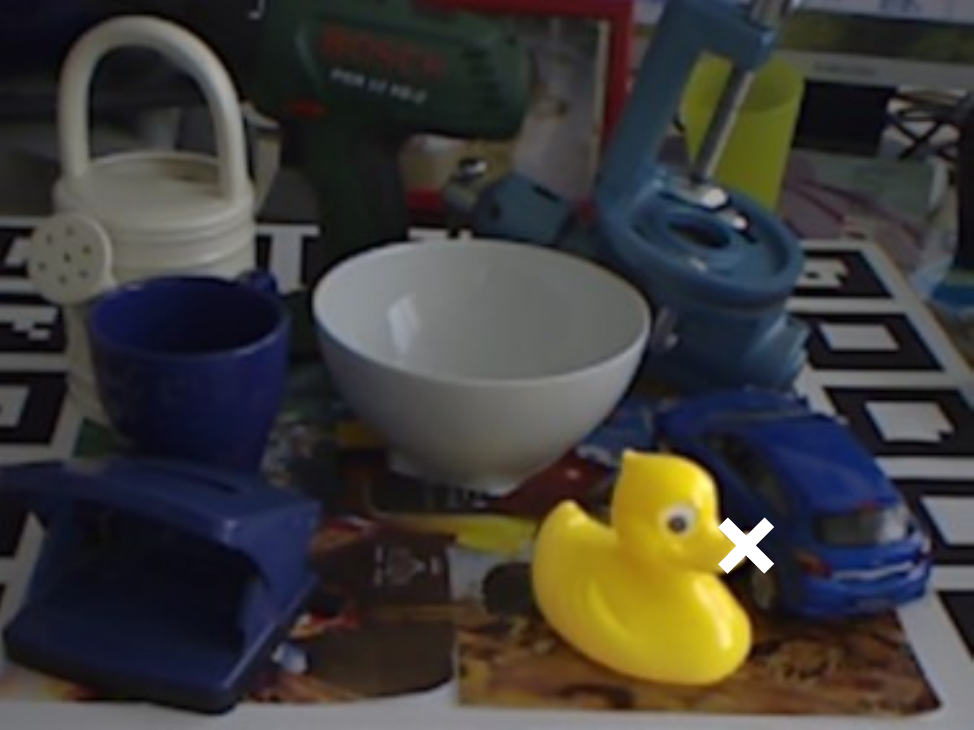} \hspace{0.25ex} &
			\includegraphics[width=0.492\linewidth]{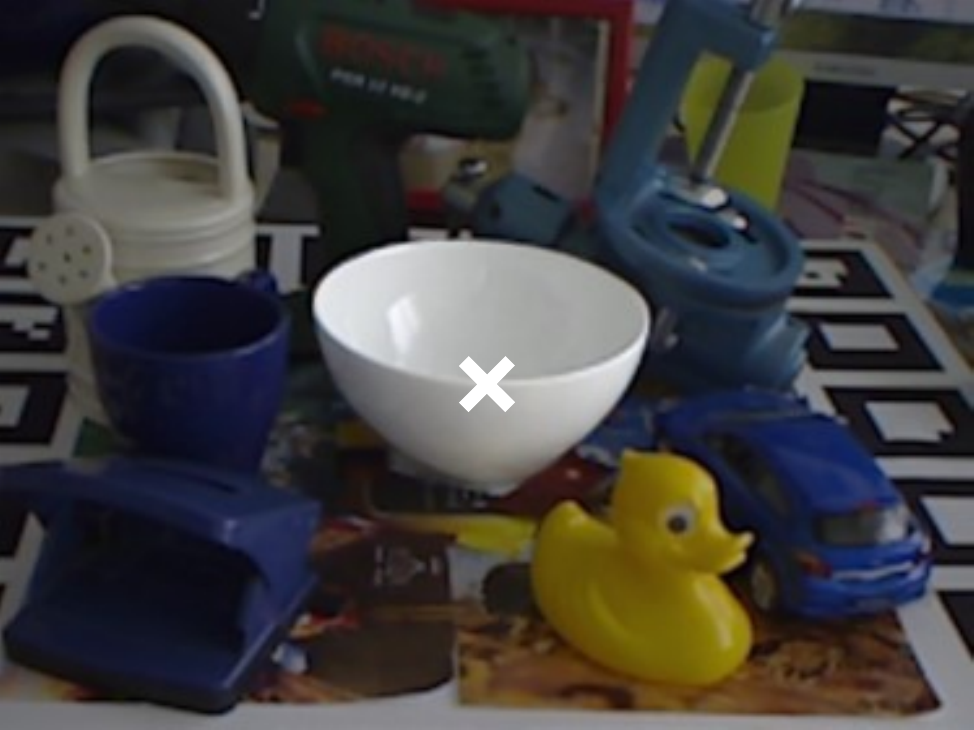} \\
		\end{tabular}
		\endgroup
		\caption{\label{fig:teaser}
			A 2D image location corresponds to a \emph{single} 3D location on the object model in the case of distinct object parts (left), but to \emph{multiple} 3D locations in the case of global or partial object symmetries (right). Representing an object by surface fragments allows predicting \emph{possibly multiple} correspondences per pixel. \vspace{0.7em}
		}
	\end{center}
\end{figure}

Establishing 2D-3D correspondences is challenging for objects with global or partial symmetries~\cite{mitra2006partial} in shape or texture.
The visible part of such objects, which is determined by self-occlusions and occlusions by other objects, may have multiple fits to the object model.
Consequently, the corresponding 2D and 3D locations form a~\emph{many-to-many} relationship, \ie a 2D image location may correspond to multiple 3D locations on the model surface (Fig.~\ref{fig:teaser}), and vice versa.
This degrades the performance of methods assuming a one-to-one relationship.
Additionally, methods~relying on local image features have a poor performance on texture-less objects, because the feature detectors often fail to provide a sufficient number of reliable locations and the descriptors are no longer discriminative enough~\cite{tombari2013bold,hodan2015detection}.

This work proposes a method for estimating 6D pose of possibly multiple instances of possibly multiple rigid objects with available 3D models from a single RGB input image.
The method is applicable to a broad range of objects -- besides those with distinct and non-repeatable shape or texture (a shoe, box of corn flakes,~\etc \cite{lowe1999object,collet2011moped}), the method handles texture-less objects and objects with global or partial symmetries (a~bowl, cup,~\etc \cite{hodan2017tless,drost2017introducing,hinterstoisser2012accv}).

\begin{figure}[t!]
	\begin{center}
		\includegraphics[width=1.0\linewidth]{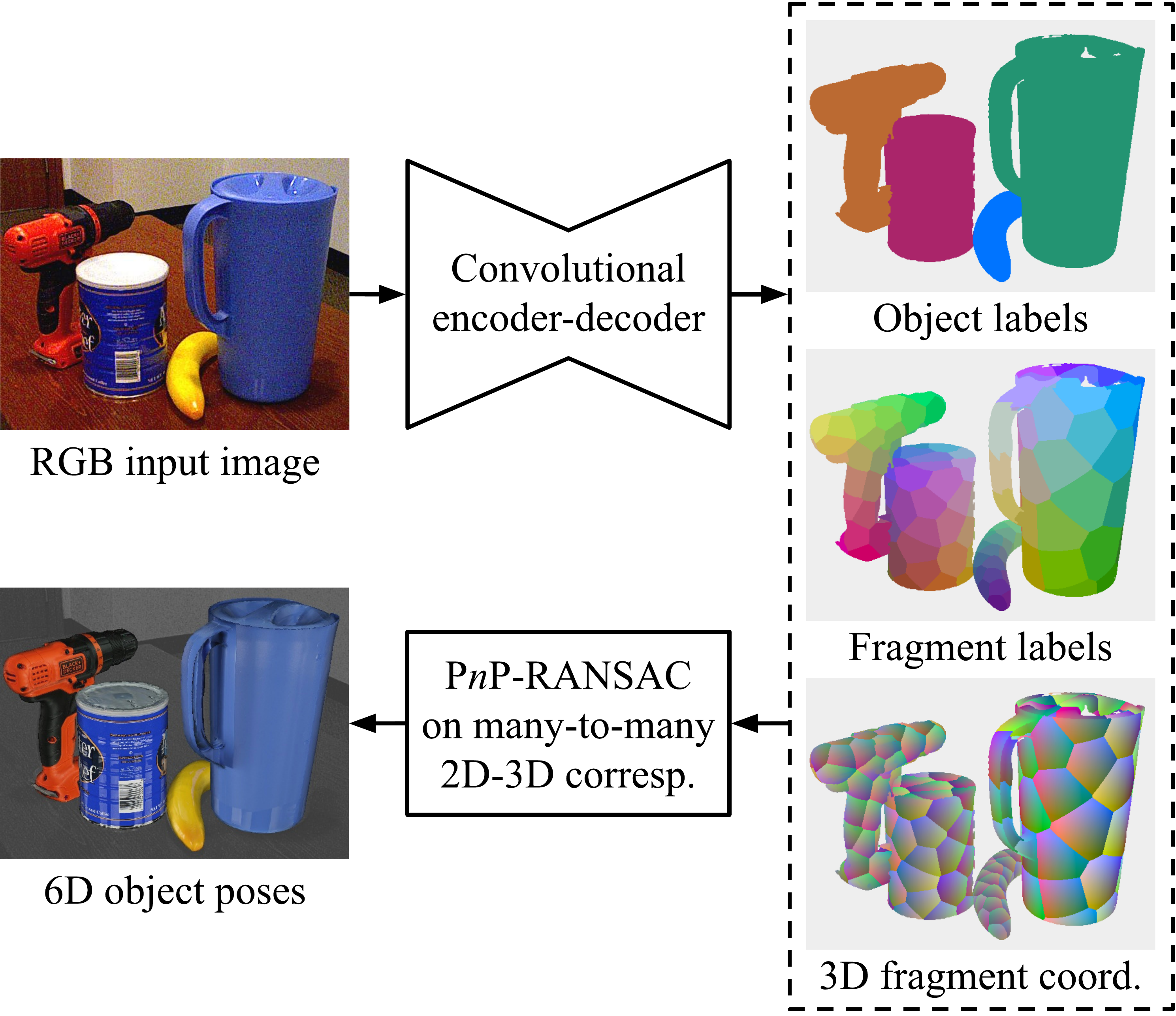}\vspace{0.5ex}
		\caption{\label{fig:pipeline} \textbf{EPOS pipeline.}
			During training, an encoder-decoder network is provided a per-pixel annotation in the form of an object label, a fragment label, and 3D fragment coordinates.
			During inference, 3D locations on \emph{possibly multiple} fragments are predicted at each pixel, which allows to capture object symmetries.
			Many-to-many 2D-3D correspondences are established by linking pixels with the predicted 3D locations, and a robust and efficient variant of the P\emph{n}P-RANSAC algorithm is used to estimate the 6D poses.
		}
	\end{center}
\end{figure}

The key idea is to represent an object by a controllable number of compact surface fragments.
This representation allows handling symmetries in a systematic manner and ensures a consistent number and uniform coverage of candidate 3D locations on objects of any type.
Correspondences between densely sampled pixels and the surface fragments are predicted using an encoder-decoder convolutional neural network.
At each pixel, the network predicts (i) the probability of each object's presence, (ii) the probability of the fragments given the object's presence, and (iii) the precise 3D location on each fragment (Fig.~\ref{fig:pipeline}).
By modeling the probability of fragments conditionally, the uncertainty due to object symmetries is decoupled from the uncertainty of the object's presence and is used to guide the selection of a data-dependent number of 3D locations at each pixel.

Poses of possibly multiple object instances are estimated from the predicted many-to-many 2D-3D correspondences by a robust and efficient variant of the P\emph{n}P-RANSAC algorithm~\cite{lepetit2009epnp} integrated in the Progressive-X scheme~\cite{barath2019progx}.
Pose hypotheses are proposed by GC-RANSAC~\cite{barath2018gcransac} which utilizes the spatial coherence of correspondences -- close correspondences (in 2D and 3D) likely belong to the same pose.
Efficiency is achieved by the PROSAC sampler~\cite{chum2005matching} that prioritizes correspondences with a high predicted probability.

The proposed method is compared with the participants of the BOP Challenge 2019~\cite{bop19challenge,hodan2018bop}.
The method outperforms all RGB methods and most RGB-D and D methods on the T-LESS~\cite{hodan2017tless} and LM-O~\cite{brachmann2014learning} datasets, which include texture-less and symmetric objects captured in cluttered scenes under various levels of occlusion.
On the YCB-V~\cite{xiang2017posecnn} dataset, which includes textured and texture-less objects, the method is superior to all competitors, with a significant $27\%$ absolute improvement over the second-best RGB method.
These results are achieved without any post-refinement of the estimated poses, such as~\cite{manhardt2018deep,li2018deepim,zakharov2019dpod,rad2017bb8}.

\vspace{1.643ex}
\noindent
This work makes the following contributions:
\vspace{-1.0ex}

\begin{enumerate}[topsep=0pt,itemsep=-1ex,partopsep=1ex,parsep=1ex]
\item \emph{A 6D object pose estimation method} applicable to a broad range of objects, including objects with symmetries, achieving the state-of-the-art RGB-only results on the standard T-LESS, YCB-V and LM-O datasets.
\item \emph{Object representation by surface fragments} allowing to handle symmetries in a systematic manner and ensuring a consistent number and uniform coverage of candidate 3D locations on any object.
\item \emph{Many-to-many 2D-3D correspondences} established by predicting a data-dependent number of precise 3D locations at each pixel.
\item  \emph{A robust and efficient estimator} for recovering poses of multiple object instances, with a demonstrated benefit over standard P\emph{n}P-RANSAC variants.
\end{enumerate}

%%%%%%%%%%%%%%%%%%%%%%%%%%%%%%%%%%%%%%%%%%%%%%%%%%%%%%%%%%%%%%%%%%%%%%%%%%%%%%%%
\section{Related Work}

\noindent
\textbf{Classical Methods.}
In the early attempt, Roberts~\cite{roberts1963machine} assumed that objects can be constructed from transformations of known simple 3D models which were fit to edges extracted from a grayscale input image.
The first practical approaches were relying on local image features~\cite{lowe1999object,collet2011moped} or template matching~\cite{brunelli2009template}, and assumed a grayscale or RGB input image.
Later, with the introduction of the consumer-grade Kinect-like sensors, the attention of the research field was steered towards estimating the object pose from RGB-D images.
Methods based on RGB-D template matching~\cite{hinterstoisser2012accv,hodan2015detection}, point-pair features~\cite{drost2010model,hinterstoisser2016going,vidal2018method}, 3D local features~\cite{guo2016comprehensive}, and learning-based methods~\cite{brachmann2014learning,tejani2014latent,krull2015learning} demonstrated a superior performance over RGB-only counterparts.

\customparagraph{CNN-Based Methods.}
Recent methods are based on convolutional neural networks (CNN's) and focus primarily on estimating the object pose from RGB images.
A popular approach is to establish 2D-3D correspondences by predicting the 2D projections of a fixed set of 3D keypoints, which are pre-selected for each object model, and solve for the object pose using P\emph{n}P-RANSAC~\cite{rad2017bb8,pavlakos20176,oberweger2018making,tekin2018real,tremblay2018deep,fu2019deephmap++,hu2019segmentation,peng2019pvnet}.
Methods establishing the correspondences in the opposite direction, \ie~by predicting the 3D object coordinates~\cite{brachmann2014learning} for a densely sampled set of pixels, have been also proposed~\cite{jafari2018ipose,nigam2018detect,zakharov2019dpod,park2019pix2pose,li2019cdpn}.
As discussed below, none of the existing correspondence-based methods can reliably handle pose ambiguity due to object symmetries.

Another approach is to localize the objects with 2D bounding boxes, and for each box predict the pose by regression~\cite{xiang2017posecnn,li2018unified,manhardt2019explaining} or by classification into discrete viewpoints~\cite{kehl2017ssd,corona2018pose,sundermeyer2019augmented}.
However, in the case of occlusion, estimating accurate 2D bounding boxes covering the whole object (including the invisible parts) is problematic~\cite{kehl2017ssd}.

Despite promising results, the recent CNN-based RGB methods are inferior to the classical RGB-D and D methods, as reported in~\cite{bop19challenge,hodan2018bop}.
Using the depth image as an additional input of the CNN is a promising research direction~\cite{li2018unified,sock2018multi,wang2019densefusion}, but with a limited range of applications.

\customparagraph{Handling Object Symmetries.}
The many-to-many relationship of corresponding 2D and 3D locations, which arises in the case of object symmetries (Sec.~\ref{sec:intro}), degrades the performance of correspondence-based methods which assume a one-to-one relationship.
In particular, classification-based methods predict for each pixel up to one corresponding 3D location~\cite{brachmann2014learning,nigam2018detect}, or for each 3D keypoint up to one 2D location which is typically given by the maximum response in a predicted heatmap~\cite{pavlakos20176,oberweger2018making,fu2019deephmap++}.
This may result in a set of correspondences which carries only a limited support for each of the possible poses.
On the other hand, regression-based methods~\cite{tekin2018real,zakharov2019dpod,peng2019pvnet} need to compromise among the possible corresponding locations and tend to return the average, which is often not a valid solution.
For example, the average of all points on a sphere is the center of the sphere, which is not a valid surface location.

The problem of pose ambiguity due to object symmetries has been approached by several methods.
Rad and Lepetit~\cite{rad2017bb8} assume that the global object symmetries are known and propose a pose normalization applicable to the case when the projection of the axis of symmetry is close to vertical.
Pitteri~\etal~\cite{pitteri2019object} introduce a pose normalization that is not limited to this special case.
Kehl~\etal~\cite{kehl2017ssd} train a classifier for only a subset of viewpoints defined by global object symmetries.
Corona~\etal~\cite{corona2018pose} show that predicting the order of rotational symmetry can improve the accuracy of pose estimation.
Xiang~\etal~\cite{xiang2017posecnn} optimize a loss function that is invariant to global object symmetries.
Park~\etal~\cite{park2019pix2pose} guide pose regression by calculating the loss \wrt to the closest symmetric pose.
However, all of these approaches cover only pose ambiguities due to global object symmetries. Ambiguities due to partial object symmetries (\ie when the visible object part has multiple possible fits to the entire object surface) are not covered.

As EPOS, the methods by Manhardt~\etal~\cite{manhardt2019explaining} and Li~\etal~\cite{li2018unified} can handle pose ambiguities due to both global and partial object symmetries without requiring any a priori information about the symmetries.
The first~\cite{manhardt2019explaining} predicts multiple poses for each object instance to estimate the distribution of possible poses induced by symmetries.
The second~\cite{li2018unified} deals with the possibly non-unimodal pose distribution by a classification and regression scheme applied to the rotation and translation space.
Nevertheless, both methods rely on estimating accurate 2D bounding boxes which is problematic when the objects are occluded~\cite{kehl2017ssd}.

\customparagraph{Object Representation.}
To increase the robustness of 6D object pose tracking against occlusion, Crivellaro~\etal~\cite{crivellaro2017robust} represent an object by a set of parts and estimate the 6D pose of each part by predicting the 2D projections of pre-selected 3D keypoints.
Brachmann~\etal~\cite{brachmann2014learning} and Nigam~\etal~\cite{nigam2018detect} split the 3D bounding box of the object model into uniform bins and predict up to one corresponding bin per pixel.
They represent each bin with its center which yields correspondences with limited precision.

For human pose estimation, G\"{u}ler~\etal~\cite{alp2018densepose} segment the 3D surface of the human body into
semantically-defined parts.
At each pixel, they predict a label of the corresponding part and the UV texture coordinates defining the precise location on the part.
In contrast, to effectively capture the partial object symmetries, we represent an object by a set of compact surface fragments of near-uniform size and predict possibly multiple labels of the corresponding fragments per pixel.
Besides, we regress the precise location in local 3D coordinates of the fragment instead of the UV coordinates.
Using the UV coordinates requires a well-defined topology of the mesh model, which may need manual intervention, and is problematic for objects with a complicated surface such as a coil or an engine~\cite{drost2017introducing}.

\customparagraph{Model Fitting.}
Many of the recent correspondence-based methods, \eg~\cite{park2019pix2pose,zakharov2019dpod,rad2017bb8,tekin2018real}, estimate the pose using the vanilla P\emph{n}P-RANSAC algorithm~\cite{fischler1981random,lepetit2009epnp} implemented in the OpenCV function \texttt{solvePnPRansac}.
We show that a noticeable improvement can be achieved by replacing the vanilla with a modern robust estimator.

%%%%%%%%%%%%%%%%%%%%%%%%%%%%%%%%%%%%%%%%%%%%%%%%%%%%%%%%%%%%%%%%%%%%%%%%%%%%%%%%
\section{EPOS: The Proposed Method} \label{sec:method}

This section provides a detailed description of the proposed model-based method for 6D object pose estimation.
The 3D object models are the only necessary training input of the method.
Besides a synthesis of automatically annotated training images~\cite{hodan2019photorealistic}, the models are useful for applications such as robotic grasping or augmented reality.

\subsection{Surface Fragments} \label{sec:surface_seg}

A mesh model defined by a set of 3D vertices, $V_i$, and a set of triangular faces, $T_i$, is assumed available for each object with index $i \in I = \{1, \dots, m\}$.
The set of all 3D points on the model surface, $S_i$, is split into $n$ fragments with indices $J = \{1, \dots, n\}$.
Surface fragment $j$ of object $i$ is defined as $S_{ij} = \{\boldsymbol{x} \, | \, \boldsymbol{x} \in S_i \, \land \, d(\boldsymbol{x}, \boldsymbol{g}_{ij}) < d(\boldsymbol{x}, \boldsymbol{g}_{ik})\}$, $\forall k \in J, k \neq j$, where $d(.)$ is the Euclidean distance of two 3D points and $\{\boldsymbol{g}_{ij}\}_{j=1}^n$ are pre-selected fragment centers.

The fragment centers are found by the furthest point sampling algorithm which iteratively selects the vertex from $V_i$ that is furthest from the already selected vertices. The algorithm starts with the centroid of the object model which is then discarded from the final set of centers.

\subsection{Prediction of 2D-3D Correspondences}
\label{sec:coord_pred}
\noindent
\textbf{Decoupling Uncertainty Due to Symmetries.}
The probability of surface fragment $j$ of object $i$ being visible at pixel $\boldsymbol{u} = (u, v)$ is modeled as:
\begin{align*}
\Pr(f\myeq j,o\myeq i\,|\,\boldsymbol{u}) = \Pr(f\myeq j\,|\,o\myeq i,\boldsymbol{u})\Pr(o\myeq i\,|\,\boldsymbol{u}),
\end{align*}
where $o$ and $f$ are random variables representing the object and fragment respectively.
The probability can be low because (1) object $i$ is not visible at pixel $\boldsymbol{u}$, or (2) $\boldsymbol{u}$ corresponds to multiple fragments due to global or partial symmetries of object $i$.
To disentangle the two cases, we predict $a_{i}(\boldsymbol{u}) = \Pr(o\myeq i\,|\,\boldsymbol{u})$ and $b_{ij}(\boldsymbol{u}) = \Pr(f\myeq j\,|\,o\myeq i,\boldsymbol{u})$ separately, instead of directly predicting $\Pr(f\myeq j,o\myeq i\,|\,\boldsymbol{u})$.

\customparagraph{Regressing Precise 3D Locations.}
Surface fragment~$j$ of object~$i$ is associated with a regressor, $\boldsymbol{r}_{ij}: \mathbb{R}^2 \rightarrow \mathbb{R}^3$, which at pixel~$\boldsymbol{u}$ predicts the corresponding 3D location: $\boldsymbol{r}_{ij}(\boldsymbol{u}) = (\boldsymbol{x} - \boldsymbol{g}_{ij}) / h_{ij}$.\ \hspace{0.18ex}The predicted location is expressed in \emph{3D~fragment coordinates}, \ie in a 3D coordinate frame with the origin at the fragment center $\boldsymbol{g}_{ij}$. Scalar $h_{ij}$ normalizes the regression range and is defined as the length of the longest side of the 3D bounding box of the fragment.

\customparagraph{Dense Prediction.}
A single deep convolutional neural network with an encoder-decoder structure, DeepLabv3+~\cite{chen2018encoder}, is adopted to densely predict $a_{k}(\boldsymbol{u})$, $b_{ij}(\boldsymbol{u})$ and $\boldsymbol{r}_{ij}(\boldsymbol{u})$, $\forall i \in I$, $\forall j \in J$, $\forall k \in I \cup \{0\}$, where $0$ is reserved for the background class.
For $m$ objects, each represented by $n$ surface fragments, the network has $4mn$$+$$m$$+$$1$ output channels ($m$$+$$1$ for probabilities of the objects and the background, $mn$ for probabilities of the surface fragments, and $3mn$ for the 3D fragment coordinates).

\customparagraph{Network Training.}
The network is trained by minimizing the following loss averaged over all pixels $\boldsymbol{u}$:
\begin{align*}
\begin{split}
L(\boldsymbol{u}) = \, &
E\big(\bar{\boldsymbol{a}}{(\boldsymbol{u})}, \boldsymbol{a}{(\boldsymbol{u})}\big) + \\
&\sum\nolimits_{i\in I} \bar{a}_{i}{(\boldsymbol{u})}
\Big[\lambda_1 E\big(\bar{\boldsymbol{b}}_{i}{(\boldsymbol{u})}, \boldsymbol{b}_{i}{(\boldsymbol{u})}\big) + \\
&\sum\nolimits_{j \in J}  \bar{b}_{ij}{(\boldsymbol{u})}\lambda_2 H\big(\bar{\boldsymbol{r}}_{ij}{(\boldsymbol{u})}, \boldsymbol{r}_{ij}{(\boldsymbol{u})}\big)\Big],
\end{split}
\end{align*}
where $E$ is the softmax cross entropy loss and $H$ is the Huber loss~\cite{huber1992robust}.
Vector $\boldsymbol{a}(\boldsymbol{u})$ consists of all predicted probabilities $a_i(\boldsymbol{u})$, and vector $\boldsymbol{b}_{i}(\boldsymbol{u})$ of all predicted probabilities $b_{ij}(\boldsymbol{u})$ for object $i$.
The ground-truth one-hot vectors $\bar{\boldsymbol{a}}{(\boldsymbol{u})}$ and $\bar{\boldsymbol{b}}_{i}{(\boldsymbol{u})}$ indicate which object (or the background) and which fragment is visible at~$\boldsymbol{u}$. Elements of these ground-truth vectors are denoted as $\bar{a}_{i}{(\boldsymbol{u})}$ and $\bar{b}_{ij}{(\boldsymbol{u})}$. Vector $\bar{\boldsymbol{b}}_{i}{(\boldsymbol{u})}$ is defined only if object $i$ is present at $\boldsymbol{u}$.
The ground-truth 3D fragment coordinates are denoted as $\bar{\boldsymbol{r}}_{ij}(\boldsymbol{u})$. Weights $\lambda_1$ and $\lambda_2$ are used to balance the loss terms.

The network is trained on images annotated with ground-truth 6D object poses.
Vectors $\bar{\boldsymbol{a}}(\boldsymbol{u})$, $\bar{\boldsymbol{b}}_{i}(\boldsymbol{u})$, and $\bar{\boldsymbol{r}}_{ij}(\boldsymbol{u})$ are obtained by rendering the 3D object models in the ground-truth poses with a custom OpenGL shader.
Pixels outside the visibility masks of the objects are considered to be the background.
The masks are calculated as in~\cite{hodan2016evaluation}.

\customparagraph{Learning Object Symmetries.}
Identifying all possible correspondences for training the network is not trivial.
One would need to identify the visible object parts in each training image and find their fits to the object models.
Instead, we provide the network with only a single corresponding fragment per pixel during training
and let the network learn the object symmetries implicitly.
Minimizing the softmax cross entropy loss $E\left(\bar{\boldsymbol{b}}_i(\boldsymbol{u}), \boldsymbol{b}_i(\boldsymbol{u})\right)$ corresponds exactly to minimizing the Kullback-Leibler divergence of distributions $\bar{\boldsymbol{b}}_i(\boldsymbol{u})$ and $\boldsymbol{b}_i(\boldsymbol{u})$~\cite{goodfellow2016deep}.
Hence, if the ground-truth one-hot distribution $\bar{\boldsymbol{b}}_i(\boldsymbol{u})$ indicates a different fragment at pixels with similar appearance, the network is expected to learn at such pixels the same probability $b_{ij}(\boldsymbol{u})$ for all the indicated fragments.
This assumes that the object poses are distributed uniformly in the training images, which is easy to ensure with synthetic training images.

\customparagraph{Establishing Correspondences.}
Pixel~$\boldsymbol{u}$ is linked with a 3D location, $\boldsymbol{x}_{ij}(\boldsymbol{u}) = h_{ij} \boldsymbol{r}_{ij}(\boldsymbol{u}) + \boldsymbol{g}_{ij}$, on every fragment for which $a_i(\boldsymbol{u}) > \tau_a$ and $b_{ij}(\boldsymbol{u})/\max_{k=1}^n\!\left(b_{ik}(\boldsymbol{u})\right) > \tau_b$.
Threshold $\tau_b$ is relative to the maximum to collect locations from all indistinguishable fragments that are expected to have similarly high probability~$b_{ij}(\boldsymbol{u})$.
For example, the probability distribution on a sphere is expected to be uniform, \ie $b_{ij}(\boldsymbol{u})=1/n, \forall j \in J$.
On a bowl, the probability is expected to be constant around the axis of symmetry.

The set of correspondences established for instances of object~$i$ is denoted as $C_i = \left\{\left(\boldsymbol{u}, \boldsymbol{x}_{ij}(\boldsymbol{u}), s_{ij}(\boldsymbol{u})\right)\right\}$, where $s_{ij}(\boldsymbol{u}) = a_i(\boldsymbol{u})b_{ij}(\boldsymbol{u})$ is the confidence of a correspondence.
The set forms a many-to-many relationship between the 2D image locations and the predicted 3D locations.

\subsection{Robust and Efficient 6D Pose Fitting}
\vspace{-1.5ex}
\customparagraph{Sources of Outliers.}
With respect to a single object pose hypothesis, set $C_i$ of the many-to-many 2D-3D correspondences includes three types of outliers. 
First, it includes outliers due to erroneous prediction of the 3D locations.
Second, for each 2D/3D location there is up to one correspondence which is compatible with the pose hypothesis; the other correspondences act as outliers.
Third, correspondences originating from different instances of object $i$ are also incompatible with the pose hypothesis.
Set $C_i$ may be therefore contaminated with a high proportion of outliers and a robust estimator is needed to achieve stable results.

\customparagraph{Multi-Instance Fitting.} \label{sec:fitting}
To estimate poses of possibly multiple instances of object $i$ from correspondences $C_i$, we use a robust and efficient variant of the P\emph{n}P-RANSAC algorithm~\cite{fischler1981random,lepetit2009epnp} integrated in the Progressive-X scheme~\cite{barath2019progx}\footnote{\url{https://github.com/danini/progressive-x}}.
In this scheme, pose hypotheses are proposed sequentially and added to a set of maintained hypotheses by the PEARL optimization~\cite{isack2012energy}, which minimizes the energy calculated over all hypotheses and correspondences.
PEARL utilizes the spatial coherence of correspondences -- the closer they are (in 2D and 3D), the more likely they belong to the same pose of the same object instance.
To reason about the spatial coherence, a neighborhood graph is constructed by describing each correspondence by a 5D vector consisting of the 2D and 3D coordinates (in pixels and centimeters), and linking two 5D descriptors if their Euclidean distance is below threshold $\tau_d$.
The inlier-outlier threshold, denoted as $\tau_r$, is set manually and defined on the re-projection error~\cite{lepetit2009epnp}.

\customparagraph{Hypothesis Proposal.}
The pose hypotheses are proposed by GC-RANSAC~\cite{barath2018gcransac}\footnote{\url{https://github.com/danini/graph-cut-ransac}}, a locally optimized RANSAC which selects the inliers~by the $s$-$t$ graph-cut optimization.
GC-RANSAC utilizes the spatial coherence via the same neighborhood graph as PEARL.
The pose is estimated from a sampled triplet of correspondences by the P3P solver~\cite{kneip2011novel}, and refined from all inliers by the EP\emph{n}P solver~\cite{lepetit2009epnp} followed by the Levenberg-Marquardt optimization~\cite{more1978levenberg}.
The triplets are sampled by PROSAC~\cite{chum2005matching}, which first focuses on correspondences with high confidence~$s_{ij}$ (Sec.~\ref{sec:coord_pred}) and progressively blends to a uniform sampling.

\customparagraph{Hypothesis Verification.}
Inside GC-RANSAC, the quality of a pose hypothesis, denoted as $\hat{\textbf{P}}$, is calculated as:
\begin{align*}
q = 1 / |U_i| \sum\nolimits_{\boldsymbol{u} \in U_i} \max\nolimits_{\boldsymbol{c} \in C_{i\boldsymbol{u}}} \max\!\Big(0, \, 1 - e^2\big(\hat{\textbf{P}}, \boldsymbol{c}\big) / \tau_r^2\Big),
\end{align*}
where $U_i$ is a set of pixels at which correspondences $C_i$ are established, $C_{i\boldsymbol{u}} \subset C_i$ is a subset established at pixel $\boldsymbol{u}$, $e\big(\hat{\textbf{P}}, \boldsymbol{c}\big)$ is the re-projection error~\cite{lepetit2009epnp}, and $\tau_r$ is the inlier-outlier threshold.
At each pixel, quality $q$ considers only the most accurate correspondence as only up to one correspondence may be compatible with the hypothesis; the others provide alternative explanations and should not influence the quality.
GC-RANSAC runs for up to $\tau_i$ iterations until quality $q$ of an hypothesis reaches threshold $\tau_q$.
The hypothesis with the highest $q$ is the outcome of each proposal stage and is integrated into the set of maintained hypotheses.

\customparagraph{Degeneracy Testing.}
Sampled triplets which form 2D triangles with the area below $\tau_t$ or have collinear 3D locations are rejected. Moreover, pose hypotheses behind the camera or with the determinant of the rotation matrix equal to $-1$ (\ie~an improper rotation matrix~\cite{haber2011three}) are discarded.

\input{src/figure_qualitative}

%%%%%%%%%%%%%%%%%%%%%%%%%%%%%%%%%%%%%%%%%%%%%%%%%%%%%%%%%%%%%%%%%%%%%%%%%%%%%%%%
\section{Experiments}~\label{sec:experiments}
This section compares the performance of EPOS with other model-based methods for 6D object pose estimation and presents ablation experiments.

\subsection{Experimental Setup}

\noindent
\textbf{Evaluation Protocol.}
We follow the evaluation protocol of the BOP Challenge 2019~\cite{bop19challenge,hodan2018bop} (BOP19 for short).
The task is to estimate the 6D poses of a varying number of instances of a varying number of objects in a single image, with the number of instances provided with each image.

The error of an estimated pose $\hat{\textbf{P}}$ \wrt the ground-truth pose $\bar{\textbf{P}}$ is calculated by three pose-error functions. The first, Visible Surface Discrepancy, treats indistinguishable poses as equivalent by considering only the visible object part:
\begin{align*}
&e_\mathrm{VSD} =
\underset{p \in \hat{V} \cup \bar{V}}{\mathrm{avg}}
\begin{cases}
0 & \text{if} \, p \in \hat{V} \cap \bar{V} \, \wedge \, |\hat{D}(p) -
\bar{D}(p)| < \tau \\
1 & \text{otherwise},
\end{cases}
\end{align*}
where $\hat{D}$ and $\bar{D}$ are distance maps obtained by rendering the object model in the estimated and the ground-truth pose respectively.
The distance maps are compared with distance map $D_I$ of test image $I$ in order to obtain visibility masks $\hat{V}$ and~$\bar{V}$, \ie~sets of pixels where the object model is visible in image~$I$.
The distance map $D_I$ is available for all images included in BOP.
Parameter $\tau$ is a misalignment tolerance.

The second pose-error function, Maximum Symmetry-Aware Surface Distance, measures the surface deviation in 3D and is therefore relevant for robotic applications:
\begin{align*}
&e_{\text{MSSD}} = \text{min}_{\textbf{T} \in T_i} \text{max}_{\textbf{x}
\in V_i}
\Vert \hat{\textbf{P}}\textbf{x} - \bar{\textbf{P}}\textbf{T}\textbf{x}
\Vert_2,
\end{align*}
where $T_i$ is a set of symmetry transformations of object $i$ (provided in BOP19), and $V_i$ is a set of model vertices.

The third pose-error function, Maximum Symmetry-Aware Projection Distance, measures the perceivable deviation.
It is relevant for augmented reality applications and suitable for the evaluation of RGB methods, for which estimating the $Z$ translational component is more challenging:
\begin{align*}
&e_{\text{MSPD}} = \text{min}_{\textbf{T} \in T_i} \text{max}_{\textbf{x}
\in V_i}
\Vert \text{proj}( \hat{\textbf{P}}\textbf{x} ) - \text{proj}(
\bar{\textbf{P}}\textbf{T}\textbf{x} ) \Vert_2,
\end{align*}
where $\text{proj}(.)$ denotes the 2D projection operation and the meaning of the other symbols is as in $e_{\text{MSSD}}$.

An estimated pose is considered correct \wrt pose-error function~$e$ if $e < \theta_e$, where $e \in \{e_{\text{VSD}}, e_{\text{MSSD}}, e_{\text{MSPD}}\}$ and $\theta_e$ is the threshold of correctness.
The fraction of annotated object instances, for which a correct pose is estimated, is referred to as recall.
The Average Recall \wrt function~$e$ ($\text{AR}_e$) is defined as the average of the recall rates calculated for multiple settings of threshold $\theta_e$, and also for multiple settings of the misalignment tolerance $\tau$ in the case of $e_{\text{VSD}}$.
The overall performance of a method is measured by the Average Recall: $\text{AR} = (\text{AR}_{\text{VSD}} + \text{AR}_{\text{MSSD}} + \text{AR}_{\text{MSPD}}) \, / \, 3$.
As EPOS uses only RGB, besides $\text{AR}$ we report $\text{AR}_{\text{MSPD}}$.

\customparagraph{Datasets.}
The experiments are conducted on three datasets: T-LESS~\cite{hodan2017tless}, YCB-V~\cite{xiang2017posecnn}, LM-O~\cite{brachmann2014learning}.
The datasets include color 3D object models and RGB-D images of VGA resolution with ground-truth 6D object poses (EPOS uses only the RGB channels). 
The same subsets of test images as in BOP19 were used.
LM-O contains 200 test images with the ground truth for eight, mostly texture-less objects from LM~\cite{hinterstoisser2012accv} captured in a clutered scene under various levels of occlusion.
YCB-V includes 21 objects, which are both textured and texture-less, and 900 test images showing the objects with occasional occlusions and limited clutter.
T-LESS contains 30 objects with no significant texture or discriminative color, and with symmetries and mutual similarities in shape and/or size. It includes 1000 test images from 20 scenes with varying complexity, including challenging scenes with multiple instances of several objects and with a high amount of clutter and occlusion.

\customparagraph{Training Images.}
The network is trained on several types of synthetic images.
For T-LESS, we use 30K physically-based rendered (PBR) images from SyntheT-LESS~\cite{pitteri2019object}, 50K images of objects rendered with OpenGL on random photographs from NYU Depth V2~\cite{silberman2012indoor} (similarly to~\cite{hinterstoisser2017pre}), and 38K real images from~\cite{hodan2017tless} showing objects on black background, where we replaced the background with random photographs.
For YCB-V, we use the provided 113K real and 80K synthetic images.
For LM-O, we use 67K PBR images from~\cite{hodan2019photorealistic} (scenes 1 and 2), and 50K images of objects rendered with OpenGL on random photographs. No real images of the objects are used for training on LM-O.

\customparagraph{Optimization.}
We use the DeepLabv3+ encoder-decoder network~\cite{chen2018encoder} with Xception-65~\cite{chollet2017xception} as the backbone.
The network is pre-trained on Microsoft COCO~\cite{lin2014microsoft} and fine-tuned on the training images described above for 2M iterations.
The batch size is set to $1$, initial learning rate to $0.0001$, parameters of batch normalization are not fine-tuned and other hyper-parameters are set as in~\cite{chen2018encoder}.

To overcome the domain gap between the synthetic training and real test images, we apply the simple technique from~\cite{hinterstoisser2017pre} and freeze the ``early flow'' part of Xception-65.
For LM-O, we additionally freeze the ``middle flow'' since there are no real training images in this dataset.
The training images are augmented by randomly adjusting brightness, contrast, hue, and saturation, and by applying random Gaussian noise and blur, similarly to~\cite{hinterstoisser2017pre}.

\customparagraph{Method Parameters.}
The rates of atrous spatial pyramid pooling in the DeepLabv3+ network are set to $12$, $24$, and $36$, and the output stride to $8\,\text{px}$.
The spatial resolution of the output channels is doubled by the bilinear interpolation, \ie~locations $\boldsymbol{u}$ for which the predictions are made are at the centers of $4\times4\,\text{px}$ regions in the input image.
A single network per dataset is trained, each object is represented by $n$ = $64$ fragments (unless stated otherwise), and the other parameters are set as follows: $\lambda_1$~=~$1$, $\lambda_2$~=~$100$, $\tau_a$~=~$0.1$, $\tau_b$~=~$0.5$, $\tau_d$~=~$20$, $\tau_r$~=~$4\,\text{px}$, $\tau_i=400$, $\tau_q$~=~$0.5$, $\tau_t$~=~$100\,\text{px}$.

\input{src/table_bop}

\subsection{Main Results}

\noindent
\textbf{Accuracy.} Tab.~\ref{tab:bop_results} compares the performance of EPOS with the participants of the BOP Challenge 2019~\cite{bop19challenge,hodan2018bop}.
EPOS outperforms all RGB methods on all three datasets by a large margin in both $\text{AR}$ and $\text{AR}_{\text{MSPD}}$ scores.
On the YCB-V dataset, it achieves $27\%$ absolute improvement in both scores over the second-best RGB method and also outperforms all RGB-D and D methods.
On the T-LESS and LM-O datasets, which include symmetric and texture-less objects, EPOS achieves the overall best $\text{AR}_{\text{MSPD}}$ score.

As the BOP rules require the method parameters to be fixed across datasets, Tab.~\ref{tab:bop_results} reports scores achieved with objects from all datasets represented by $64$ fragments.
As reported in Tab.\ \ref{tab:fragments}, increasing the number of fragments from $64$ to $256$ yields in some cases additional improvements but around double image processing time.
Note that we do not perform any post-refinement of the estimated poses, such as \cite{manhardt2018deep,li2018deepim,zakharov2019dpod,rad2017bb8}, which could improve the accuracy further.

\customparagraph{Speed.}
With an unoptimized implementation, EPOS takes $0.75\,\text{s}$ per image on average (with a 6-core Intel i7-8700K CPU, 64GB RAM, and Nvidia P100 GPU).
As the other RGB methods, which are all based on convolutional neural networks, EPOS is noticeably faster than the RGB-D and D methods (Tab.~\ref{tab:bop_results}), which are slower typically due to an ICP post-processing step~\cite{rusinkiewicz2001efficient}.
The RGB methods of~\cite{sundermeyer2019augmented,zakharov2019dpod} are $3$--$4$ times faster but significantly less accurate than EPOS.
Depending on the application requirements, the trade-off between the accuracy and speed of EPOS can be controlled by, \eg, the number of surface fragments, the network size, the image resolution, the density of pixels at which the correspondences are predicted, or the maximum allowed number of GC-RANSAC iterations.

\subsection{Ablation Experiments}

\noindent
\textbf{Surface Fragments.}
The performance scores of EPOS for different numbers of surface fragments are shown in the upper half of Tab.~\ref{tab:fragments}.
With a single fragment, the method performs direct regression of the so-called 3D object coordinates~\cite{brachmann2014learning}, similarly to~\cite{jafari2018ipose,park2019pix2pose,li2019cdpn}.
The accuracy increases with the number of fragments and reaches the peak at $64$ or $256$ fragments.
On all three datasets, the peaks of both $\text{AR}$ and $\text{AR}_\text{MSPD}$ scores are $18$--$33\%$ higher than the scores achieved with the direct regression of the 3D object coordinates.
This significant improvement demonstrates the effectiveness of fragments on various types of objects, including textured, texture-less, and symmetric objects.

On T-LESS, the accuracy drops when the number of fragments is increased from $64$ to $256$.
We suspect this is because the fragments become too small (T-LESS includes smaller objects) and training of the network becomes challenging due to a lower number of examples per fragment.

The average number of correspondences increases with the number of fragments, \ie each pixel gets linked with more fragments (columns Corr.\ in Tab.~\ref{tab:fragments}).
At the same time, the average number of fitting iterations tends to decrease (columns Iter.).
This shows that the pose fitting method can benefit from knowing more possible correspondences per pixel -- GC-RANSAC finds a pose hypothesis with quality $q$ (Sec.~\ref{sec:fitting}) reaching threshold $\tau_q$ in less iterations.
However, although the average number of iterations decreases, the average image processing time tends to increase (at higher numbers of fragments) due to a higher computational cost of the network inference and of each fitting iteration.
Setting the number of fragments to $64$ provides a practical trade-off between the speed and accuracy.

\input{src/table_fragments}
\input{src/table_ransac}

\customparagraph{Regression of 3D Fragment Coordinates.}
The upper half of Tab.~\ref{tab:fragments} shows scores achieved with regressing the precise 3D locations, while the lower half shows scores achieved with the same network models but using the fragment centers (Sec.~\ref{sec:surface_seg}) instead of the regressed locations.
Without the regression, the scores increase with the number of fragments as the deviation of the fragment centers from the true corresponding 3D locations decreases.
However, the accuracy is often noticeably lower than with the regression.
With a single fragment and without the regression, all pixels are linked to the same fragment center and all samples of three correspondences are immediately rejected because they fail the non-collinearity test, hence the low processing time.

Even though the regressed 3D locations are not guaranteed to lie on the model surface, their average distance from the surface is less than $1\,\text{mm}$ (with $64$ and $256$ fragments), which is negligible compared to the object sizes.
No improvement was observed when the regressed locations were replaced by the closest points on the object model.

\customparagraph{Robust Pose Fitting.}
Tab.\ \ref{tab:ransac_variants} evaluates several methods for robust pose estimation from the 2D-3D correspondences: RANSAC~\cite{fischler1981random} from OpenCV, MSAC~\cite{torr2000mlesac}, and GC-RANSAC~\cite{barath2018gcransac}.
The methods were evaluated within the Progressive-X scheme (Sec.~\ref{sec:fitting}), with the P3P solver~\cite{kneip2011novel} to estimate the pose from a minimal sample, \ie three correspondences, and with several solvers to estimate the pose from a non-minimal sample.
In OpenCV RANSAC and MSAC, the non-minimal solver refines the pose from all inliers.
In GC-RANSAC, it is additionally used in the graph-cut-based local optimization which is applied when a new so-far-the-best pose is found.
We tested OpenCV RANSAC with all available non-minimal solvers and achieved the best scores with EP\emph{n}P~\cite{lepetit2009epnp}.
The top-performing estimation method on all datasets is GC-RANSAC with EP\emph{n}P followed by the Levenberg-Marquardt optimization~\cite{more1978levenberg} as the non-minimal solver.
Note the gap in accuracy, especially on T-LESS, between this method and OpenCV RANSAC.

%%%%%%%%%%%%%%%%%%%%%%%%%%%%%%%%%%%%%%%%%%%%%%%%%%%%%%%%%%%%%%%%%%%%%%%%%%%%%%%%
\section{Conclusion}

We have proposed a new model-based method for 6D object pose estimation from a single RGB image.
The key idea is to represent an object by compact surface fragments, predict possibly multiple corresponding 3D locations at each pixel, and solve for the pose using a robust and efficient variant of the P\emph{n}P-RANSAC algorithm.
The experimental evaluation has demonstrated the method to be applicable to a broad range of objects, including challenging objects with symmetries.
A study of object-specific numbers of fragments, which may depend on factors such as the physical object size, shape or the range of distances of the object from the camera, is left for future work.
The project website with source code is at: \texttt{\href{http://cmp.felk.cvut.cz/epos/}{cmp.felk.cvut.cz/epos}}.

\vspace{0.55em}
\begin{spacing}{0.85}
\begin{footnotesize}
\noindent This research was supported by Research Center for Informatics (project CZ.02.1.01/0.0/0.0/16\_019/0000765 funded by OP VVV), CTU student grant (SGS OHK3-019/20), and grant ``Exploring the Mathematical Foundations of Artificial Intelligence'' (2018-1.2.1-NKP-00008).
\end{footnotesize}
\end{spacing}
\normalsize

{\small
\bibliographystyle{ieee_fullname}
\bibliography{ref}
}

\end{document}

%% file: src/figure_qualitative.tex
\begin{figure}[t!]
	\begin{center}
	    \begingroup
        \setlength{\tabcolsep}{6.0pt}
        \renewcommand{\arraystretch}{0.0}
		\begin{tabular}{ @{}c@{ } @{}c@{ } }
			\includegraphics[width=0.495\linewidth]{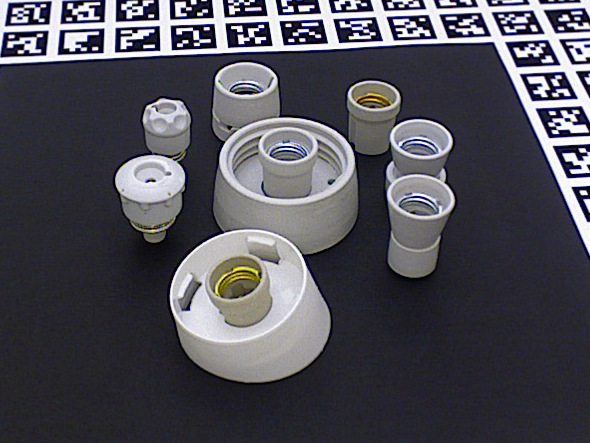} &
			\includegraphics[width=0.495\linewidth]{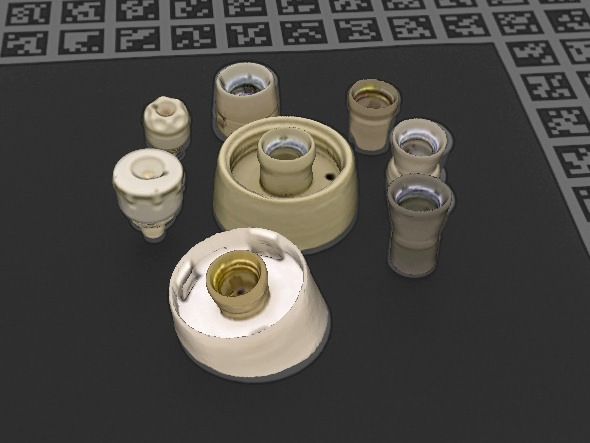}\vspace{1.25ex} \\
			
			\includegraphics[width=0.495\linewidth]{img/qualitative/ycbv/59_001070_rgb.jpg} &
			\includegraphics[width=0.495\linewidth]{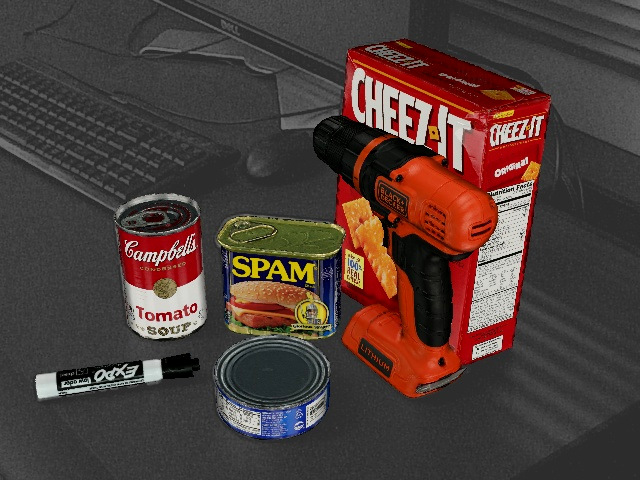}\vspace{1.25ex} \\
			
			\includegraphics[width=0.495\linewidth]{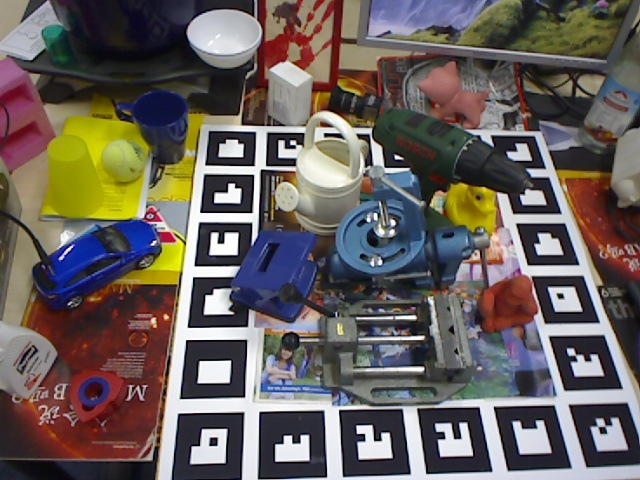} &
			\includegraphics[width=0.495\linewidth]{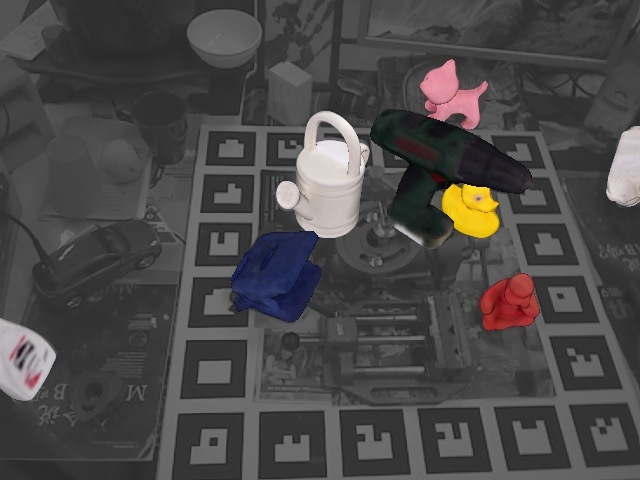}\vspace{0.5ex} \\
		\end{tabular}
		\endgroup
		\caption{\label{fig:qualitative}
			\textbf{Example EPOS results} on T-LESS (top), YCB-V (middle) and LM-O (bottom).
			On the right are renderings of the 3D object models in poses estimated from the RGB images on the left.
			All eight LM-O objects, including two truncated ones, are detected in the bottom example.
			More examples are on the project website.
		}
	\end{center}
\end{figure}

%% file: src/table_bop.tex
\setlength{\tabcolsep}{6pt}
\begin{table*}[t!]
	\begin{center}
		\begin{tabularx}{\linewidth}{l l Y Y Y Y Y Y Y}
			\toprule
			
			\multirow{2}{*}{\vspace{-0.8ex}6D object pose estimation method} &
			\multirow{2}{*}{\vspace{-0.8ex}Image} &
			\multicolumn{2}{c}{T-LESS~\cite{hodan2017tless}} &
			\multicolumn{2}{c}{YCB-V~\cite{xiang2017posecnn}} &
			\multicolumn{2}{c}{LM-O~\cite{brachmann2014learning}} &
			\multirow{2}{*}{\vspace{-0.8ex}Time} \\
			
			\cmidrule(lr){3-4} \cmidrule(lr){5-6} \cmidrule(lr){7-8}
			
			& &
			{\normalsize $\text{AR}$} &
			{\normalsize $\text{AR}_\text{{\tiny MSPD}}$} &
			{\normalsize $\text{AR}$} &
			{\normalsize $\text{AR}_\text{{\tiny MSPD}}$} &
			{\normalsize $\text{AR}$} &
			{\normalsize $\text{AR}_\text{{\tiny MSPD}}$} & \\
			
			\midrule

			\Rowcolor{lightgray} EPOS & RGB & \wincat{47.6} & \win{63.5} & \win{69.6} & \win{78.3} & \wincat{44.3} & \win{65.9} & \leavevmode\hphantom{00}0.75 \\
			
			Zhigang-CDPN-ICCV19~\cite{li2019cdpn} & RGB & 12.4 & 17.0 & 42.2 & 51.2 & 37.4 & 55.8 & \leavevmode\hphantom{00}0.67 \\
			Sundermeyer-IJCV19~\cite{sundermeyer2019augmented} & RGB & 30.4 & 50.4 & 37.7 & 41.0 & 14.6 & 25.4 & \leavevmode\hphantom{00}0.19 \\
			Pix2Pose-BOP-ICCV19~\cite{park2019pix2pose} & RGB & 27.5 & 40.3 & 29.0 & 40.7 & \leavevmode\hphantom{0}7.7 & 16.5 & \leavevmode\hphantom{00}0.81 \\
			DPOD-ICCV19 (synthetic)~\cite{zakharov2019dpod} & RGB & \leavevmode\hphantom{0}8.1 & 13.9 & 22.2 & 25.6 & 16.9 & 27.8 & \leavevmode\hphantom{00}0.24 \\
			
			\cmidrule{1-9}
			
			Pix2Pose-BOP\_w/ICP-ICCV19~\cite{park2019pix2pose} & RGB-D & \leavevmode\hphantom{00.}-- & \leavevmode\hphantom{00.}-- & \wincat{67.5} & \wincat{63.0} & \leavevmode\hphantom{00.}-- & \leavevmode\hphantom{00.}-- & \leavevmode\hphantom{000.0}-- \\
			Drost-CVPR10-Edges~\cite{drost2010model} & RGB-D & \wincat{50.0} & \wincat{51.8} & 37.5 & 27.5 & \wincat{51.5} & \wincat{56.9} & 144.10 \\
			F{\'e}lix\&Neves-ICRA17-IET19~\cite{rodrigues2019deep,raposo2017using} & RGB-D & 21.2 & 21.3 & 51.0 & 38.4 & 39.4 & 43.0 & \leavevmode\hphantom{0}52.97 \\
			Sundermeyer-IJCV19+ICP~\cite{sundermeyer2019augmented} & RGB-D & 48.7 & 51.4 & 50.5 & 47.5 & 23.7 & 28.5 & \leavevmode\hphantom{00}1.10 \\
			
			\cmidrule{1-9}
			
			Vidal-Sensors18~\cite{vidal2018method} & D & \win{53.8} & \wincat{57.4} & \wincat{45.0} & \wincat{34.7} & \win{58.2} & \wincat{64.7} & \leavevmode\hphantom{00}4.93 \\
			Drost-CVPR10-3D-Only~\cite{drost2010model} & D & 44.4 & 48.0 & 34.4 & 26.3 & 52.7 & 58.1 & \leavevmode\hphantom{0}10.47 \\
			Drost-CVPR10-3D-Only-Faster~\cite{drost2010model} & D & 40.5 & 43.6 & 33.0 & 24.4 & 49.2 & 54.2 & \leavevmode\hphantom{00}2.20 \\

			\bottomrule
		\end{tabularx}
		\caption{\label{tab:bop_results} \textbf{BOP Challenge 2019}~\cite{bop19challenge,hodan2018bop} results on datasets T-LESS, YCB-V and LM-O, with objects represented by $64$ surface fragments.
		Top scores for image types are \wincat{bold}, the best overall are \win{blue}.
		The time [s] is the average image processing time averaged over the datasets.
		} \vspace{-0.26em}
	\end{center}
\end{table*}

%% file: src/table_fragments.tex
\setlength{\tabcolsep}{0pt}
\begin{table*}[!t]
	\begin{center}
		\begin{tabularx}{\textwidth}{Y Y Y Y Y Y Y Y Y Y Y Y Y Y Y Y}

		    \toprule
			
			\multirow{2}{*}{\vspace{-0.8ex}{\normalsize $n$}} &
			\multicolumn{5}{c}{{\normalsize T-LESS~\cite{hodan2017tless}}} &
			\multicolumn{5}{c}{{\normalsize YCB-V~\cite{xiang2017posecnn}}} &
			\multicolumn{5}{c}{{\normalsize LM-O~\cite{brachmann2014learning}}} \\
			
			\cmidrule(lr){2-6} \cmidrule(lr){7-11} \cmidrule(lr){12-16}
			
			&
			{\normalsize $\text{AR}$} &
			{\normalsize $\text{AR}_\text{{\tiny MSPD}}$} &
			{\normalsize $\text{Corr.}$} &
			{\normalsize $\text{Iter.}$} &
			{\normalsize $\text{Time}$} &
			{\normalsize $\text{AR}$} &
			{\normalsize $\text{AR}_\text{{\tiny MSPD}}$} &
			{\normalsize $\text{Corr.}$} &
			{\normalsize $\text{Iter.}$} &
			{\normalsize $\text{Time}$} &
			{\normalsize $\text{AR}$} &
			{\normalsize $\text{AR}_\text{{\tiny MSPD}}$} &
			{\normalsize $\text{Corr.}$} &
			{\normalsize $\text{Iter.}$} &
			{\normalsize $\text{Time}$} \\
			
		    \cmidrule{1-16}
			\multicolumn{16}{c}{{\small \hspace{10.5mm}\textit{With regression of 3D fragment coordinates}}} \\
			\cmidrule{1-16}

			\leavevmode\hphantom{00}1 & 17.2 & 30.7 & \leavevmode\hphantom{0}911 & 347 & 0.97 & 41.7 & 52.6 & 1079 & 183 & 0.56 & 26.8 & 47.5 & 237 & 111 & 0.53 \\
			\leavevmode\hphantom{00}4 & 39.5 & 57.1 & 1196 & 273 & 0.95 & 54.4 & 66.1 & 1129 & 110 & 0.52 & 33.5 & 56.0 & 267 & \leavevmode\hphantom{0}58 & 0.51 \\
			\leavevmode\hphantom{0}16 & 45.4 & 62.7 & 1301 & 246 & 0.96 & 63.2 & 72.7 & 1174 & \leavevmode\hphantom{0}71 & 0.51 & 39.3 & 61.3 & 275 & \leavevmode\hphantom{0}54 & 0.50 \\
			\leavevmode\hphantom{0}64 & \wincat{47.6} & \wincat{63.5} & 1612 & 236 & 1.18 & 69.6 & 78.3 & 1266 & \leavevmode\hphantom{0}56 & 0.57 & 44.3 & \wincat{65.9} & 330 & \leavevmode\hphantom{0}53 & 0.49 \\
			256 & 45.6 & 59.7 & 3382 & 230 & 2.99 & \wincat{71.4} & \wincat{79.8} & 1497 & \leavevmode\hphantom{0}56 & 0.94 & \wincat{46.0} & 65.4 & 457 & \leavevmode\hphantom{0}70 & 0.60 \\
			
			\cmidrule{1-16}
			\multicolumn{16}{c}{{\small \hspace{10.5mm}\textit{Without regression of 3D fragment coordinates}}} \\
			\cmidrule{1-16}
			
			\leavevmode\hphantom{00}1 & \leavevmode\hphantom{0}0.0 & \leavevmode\hphantom{0}0.0 & \leavevmode\hphantom{0}911 & 400 & 0.23 & \leavevmode\hphantom{0}0.0 & \leavevmode\hphantom{0}0.0 & 1079 & 400 & 0.17 & \leavevmode\hphantom{0}0.0 & \leavevmode\hphantom{0}0.0 & 237 & 400 & 0.24 \\
			\leavevmode\hphantom{00}4 & \leavevmode\hphantom{0}3.2 & \leavevmode\hphantom{0}8.8 & 1196 & 399 & 0.89 & \leavevmode\hphantom{0}3.0 & \leavevmode\hphantom{0}7.4 & 1129 & 400 & 0.53 & \leavevmode\hphantom{0}5.2 & 15.2 & 267 & 390 & 0.50 \\
			\leavevmode\hphantom{0}16 & 13.9 & 37.5 & 1301 & 396 & 1.02 & 16.1 & 36.4 & 1174 & 400 & 0.61 & 17.1 & 47.7 & 275 & 359 & 0.55 \\
			\leavevmode\hphantom{0}64 & 29.4 & 55.0 & 1612 & 380 & 1.35 & 41.5 & 66.6 & 1266 & 383 & 0.73 & 31.0 & 62.3 & 330 & 171 & 0.55 \\
			256 & 43.0 & 58.2 & 3382 & 299 & 2.95 & 64.5 & 77.7 & 1497 & 206 & 0.88 & 43.2 & 64.9 & 457 & \leavevmode\hphantom{0}72 & 0.58 \\

		    \bottomrule
		\end{tabularx}
		\caption{\label{tab:fragments} \textbf{Number of fragments and regression.}
		Performance scores for different numbers of surface fragments ($n$) with and without regression of the 3D fragment coordinates (the fragment centers are used in the case of no regression). 
		The table also reports the average number of correspondences established per object model in an image, the average number of GC-RANSAC iterations to fit a single pose (both are rounded to integers), and the average image processing time [s].
		}
	\end{center}
\end{table*}

%% file: src/table_ransac.tex
\setlength{\tabcolsep}{6pt}
\begin{table*}[t!]
	\begin{center}
		\begin{tabularx}{\linewidth}{l l Y Y Y Y Y Y Y}
			\toprule
			
			\multirow{2}{*}{\vspace{-0.8ex}RANSAC variant} &
			\multirow{2}{*}{\vspace{-0.8ex}Non-minimal solver} &
			\multicolumn{2}{c}{T-LESS~\cite{hodan2017tless}} &
			\multicolumn{2}{c}{YCB-V~\cite{xiang2017posecnn}} &
			\multicolumn{2}{c}{LM-O~\cite{brachmann2014learning}} &
			\multirow{2}{*}{\vspace{-0.8ex}Time} \\
			
			\cmidrule(lr){3-4} \cmidrule(lr){5-6} \cmidrule(lr){7-8}
		
			& &
			{\normalsize $\text{AR}$} &
			{\normalsize $\text{AR}_\text{{\tiny MSPD}}$} &
			{\normalsize $\text{AR}$} &
			{\normalsize $\text{AR}_\text{{\tiny MSPD}}$} &
			{\normalsize $\text{AR}$} &
			{\normalsize $\text{AR}_\text{{\tiny MSPD}}$} & \\
			
			\midrule
			
			OpenCV RANSAC & EP\emph{n}P~\cite{lepetit2009epnp} & 35.5 & 47.9 & 67.2 & 76.6 & 41.2 & 63.5 & 0.16 \\
			MSAC~\cite{torr2002bayesian} & EP\emph{n}P~\cite{lepetit2009epnp} + LM~\cite{more1978levenberg} & 44.3 & 61.0 & 63.8 & 73.7 & 39.7 & 61.7 & 0.49 \\
			GC-RANSAC~\cite{barath2018gcransac} & DLS-P\emph{n}P~\cite{hesch2011direct} & 44.3 & 59.5 & 67.5 & 76.1 & 35.6 & 53.9 & 0.53 \\
			GC-RANSAC~\cite{barath2018gcransac} & EP\emph{n}P~\cite{lepetit2009epnp} &  46.9 & 62.6 & 69.2 & 77.9 & 42.6 & 63.6 & 0.39 \\
			GC-RANSAC~\cite{barath2018gcransac} & EP\emph{n}P~\cite{lepetit2009epnp} + LM~\cite{more1978levenberg} & \wincat{47.6} & \wincat{63.5} & \wincat{69.6} & \wincat{78.3} & \wincat{44.3} & \wincat{65.9} & 0.52 \\
			\bottomrule
		\end{tabularx}
		\caption{\label{tab:ransac_variants} 
		\textbf{RANSAC variants and non-minimal solvers.}
		The P3P solver~\cite{kneip2011novel} is used to estimate the pose from a minimal sample of 2D-3D correspondences.
		The non-minimal solvers are applied when estimating the pose from a larger-than-minimal sample.
		The reported time~[s] is the average time to fit poses of all object instances in an image averaged over the datasets.
		}
	\end{center}
\end{table*}